\documentclass[sigconf, 9pt]{acmart}
\AtBeginDocument{%
  }

\setcopyright{acmcopyright}
\copyrightyear{2018}
\acmYear{2018}
\acmDOI{XXXXXXX.XXXXXXX}

\usepackage{physics}
\usepackage[noend]{algorithmic}
\usepackage{algorithm}

\newtheorem{theorem}{Theorem}[section]
\newtheorem{proposition}[theorem]{Proposition}

\newtheorem{definition}[theorem]{Definition}

\newtheorem{problem}[theorem]{Problem}
\newtheorem{remark}[theorem]{Remark}

\begin{document}

\title{Joint Differential Optimization and Verification for Certified Reinforcement Learning}

\author{Yixuan Wang}
\authornote{Both authors contributed equally to this research.}
\affiliation{%
  \institution{Northwestern University}
  \city{Evanston}
  \state{IL}
  \country{USA}
  \postcode{60201}
}
\email{wangyixu14@gmail.com}

\author{Simon Zhan}
\authornotemark[1]
\affiliation{%
  \institution{UC Berkeley}
    \city{Berkeley}
  \state{CA}
  \country{USA}}
\email{simonzhan@berkeley.edu }

\author{Zhilu Wang}
\affiliation{%
  \institution{Northwestern University}
    \city{Evanston}
  \state{IL}
  \country{USA}
}

\author{Chao Huang}
\affiliation{%
 \institution{University of Liverpool}
 \city{Liverpool}
 \country{UK}}

\author{Zhaoran Wang}
\affiliation{%
  \institution{Northwestern University}
    \city{Evanston}
  \state{IL}
  \country{USA}
}

\author{Zhuoran Yang}
\affiliation{%
 \institution{Yale University}
 \city{New Haven}
 \state{CT}
 \country{USA}}

\author{Qi Zhu}
\affiliation{%
  \institution{Northwestern University}
    \city{Evanston}
  \state{IL}
  \country{USA}
}

\renewcommand{\shortauthors}{Anonymous Authors}
\newcommand{\yixuan}[1]{{{\color{blue} #1}}}
\newcommand{\chao}[1]{{{\color{red} #1}}}

\begin{abstract}
Model-based reinforcement learning has been widely studied for controller synthesis in cyber-physical systems (CPSs). In particular, 
for safety-critical CPSs, it is important to formally certify system properties (e.g., safety, stability) under the learned RL controller. However, as existing methods typically conduct formal verification \emph{after} the controller has been learned, it is often difficult to obtain any certificate, even after many iterations between learning and verification. To address this challenge, we propose a framework that \emph{jointly conducts reinforcement learning and formal verification} by formulating and solving a novel bilevel optimization problem, which is end-to-end differentiable by the gradients from the value function and certificates formulated by linear programs and semi-definite programs. In experiments, our framework is compared with a baseline model-based stochastic value gradient (SVG) method and its extension to solve constrained Markov Decision Processes (CMDPs) for safety. 
The results demonstrate the significant advantages of our framework in finding feasible controllers with certificates, i.e., barrier functions and Lyapunov functions that formally ensure system safety and stability.
\end{abstract}





\settopmatter{printacmref=false} 
\renewcommand\footnotetextcopyrightpermission[1]{} 
\pagestyle{plain} 

\maketitle

\section{Introduction}\label{sec:intro}

Applying machine learning techniques in cyber-physical systems (CPSs) has attracted much attention. In particular, reinforcement learning (RL) has shown great promise~\cite{wang2021cocktail}, such as in robotics~\cite{koch2019reinforcement} and smart buildings~\cite{xu2020one,Wei_DAC17}, where RL trains control policy by maximizing the value function of the goal state~\cite{sutton2018reinforcement}.
However, there is still significant hesitation in applying RL to safety-critical applications~\cite{knight2002safety,Zhu_PIEEE18} of CPSs, such as in autonomous vehicles~\cite{liu2022physics, liu2023safety, Wang_DATE22}, because of the uncertain and potentially dangerous impact on system safety~\cite{wang2020energy,zhu2021safety,Zhu_ICCAD20}.
It is thus important to find RL-learned controllers that are certified, i.e., under which critical system properties such as safety and stability can be formally guaranteed.
And a common approach to guarantee these properties is to find corresponding \emph{certificates} for them, e.g., a barrier function for safety~\cite{prajna2004safety} and a Lyapunov function for stability~\cite{lyapunov1992general}.

In this work, we focus particularly on learning certified controllers with RL for CPSs that can be modeled as  ordinary differential equations (ODEs) with unknown parameters, a common scenario in practice. Traditionally, this is typically done in a two-step \emph{`open-loop'} process: 1) first, model-based reinforcement learning (MBRL) is conducted to learn the system model parameters and the controller simultaneously, and then 2) based on the identified dynamics, formal verification is performed to find certificates for various system properties by solving optimization problems.
However, with such an open-loop paradigm, it is often difficult to find any feasible certificates even after many iterations of learning and verification steps, and the failed verification results are not leveraged sufficiently in the learning of a new controller. Thus, integrating controller synthesis and certificate generation in a more holistic manner has received increasing attention recently. 

Pioneering works on control-certificate joint learning mainly focus on systems with known models, i.e., explicit models without any unknown parameters~\cite{ qin2021learning,wang2021verification, luo2021learning, tsukamoto2020neural, robey2020learning, lindemann2021learning, dawson2022corl,chang2019neural}. Those methods typically collect samples from the system space, transform the certificate conditions into loss functions, and solve them via supervised learning methods. However, they cannot be directly used to address systems with unknown parameters, and the certificates obtained in those works are often tested/validated via sampling-based approaches without being formally verified.

Moreover, for safety properties, methods that are based on solving constrained Markov Decision Processes (CMDPs) are popular in the safe RL literature~\cite{thananjeyan2021recovery,srinivasan2020learning,stooke2020responsive}. However, these methods typically try to achieve safety by restricting the expectation of the cumulative cost for the system's unsafe actions to be under a certain threshold, which can only be regarded as \emph{soft} safety constraints as the system may still enter the unsafe region.

\begin{figure}[!ht]
  \begin{center}
    \includegraphics[width=0.95\linewidth]{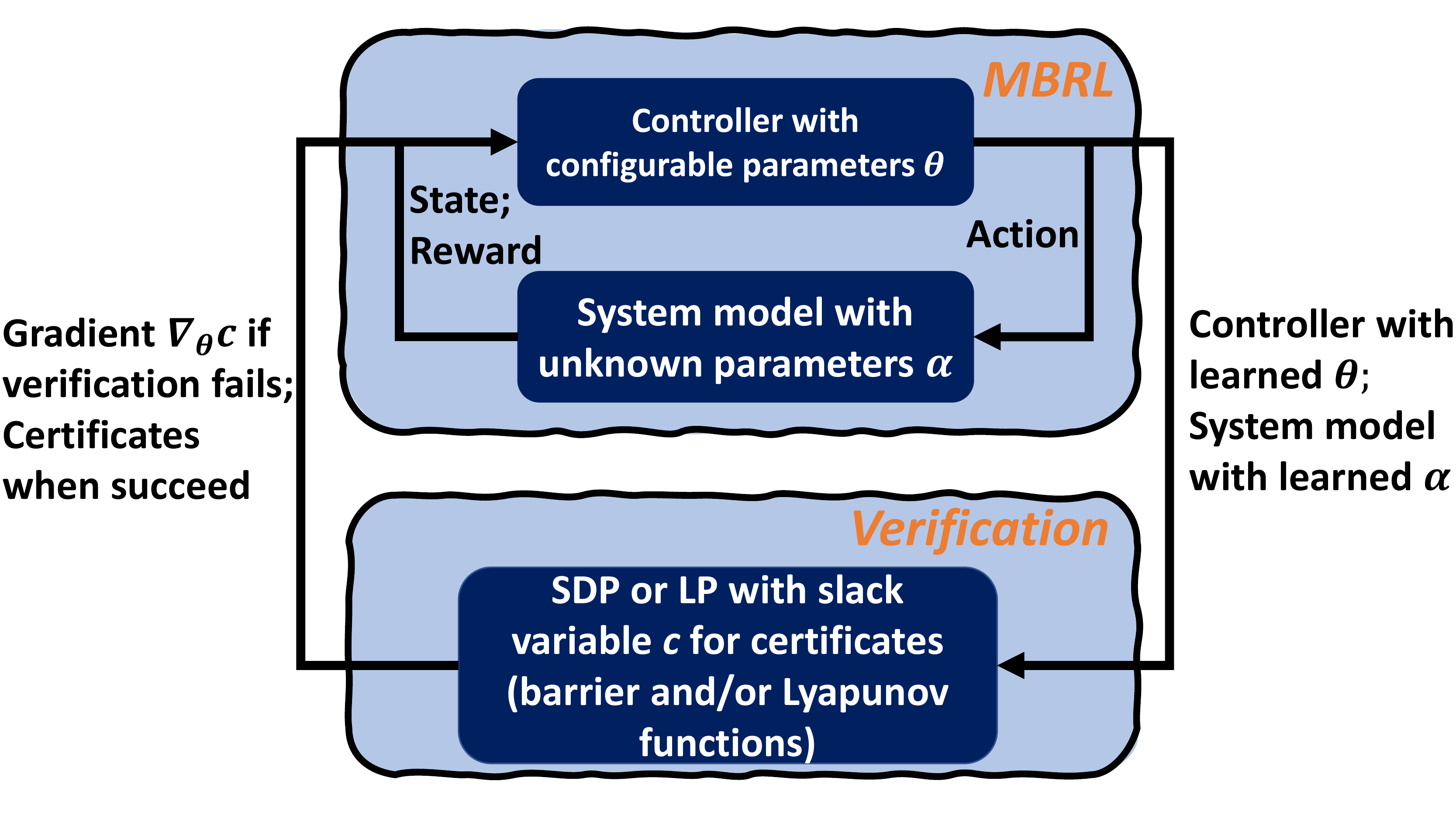}
  \end{center}
  \caption{Overview of our joint differentiable optimization and verification framework for certified RL. Our approach integrates RL-based control optimization with formal verification in a closed-loop manner, by formulating and solving a bilevel optimization problem. The upper-level problem learns the controller parameters $\theta$ and the system model parameters $\alpha$ with MBRL, while the lower-level problem verifies system properties by searching for feasible certificates via SDP or LP with a slack variable $c$. The framework is end-to-end differentiable. 
  }
   \label{fig:overview}
   \vspace{-15pt}
\end{figure}

\smallskip
\noindent
\emph{Contribution of our work:} To address these challenges, we propose a \textbf{certified RL method with joint differentiable optimization and verification for systems with unknown model parameters}. As shown in Fig.~\ref{fig:overview}, our approach seamlessly integrates RL optimization and formal verification by formulating and solving a \emph{novel bilevel optimization problem}, which generates an optimal controller together with its certificates, e.g., barrier functions for safety and/or Lyapunov functions for stability. Different from CMDP-based methods, we address \emph{hard} safety constraints where the system should never enter an unsafe region. 

The upper-level problem in our bilevel optimization tries to learn the controller parameters $\theta$ and the unknown system model parameters $\alpha$ by MBRL; while the lower-level problem tries to verify system properties by searching for feasible certificates via SDP or LP with a slack variable $c$. Note that we propose LP relaxation in addition to SDP because while SDP may be more efficient for low-dimensional/low-degree systems, LP provides better scalability for higher-dimensional systems.
When the lower-level problem fails to find any feasible certificate for given $\theta$ and $\alpha$ from the upper-level MBRL problem, it will return the gradient of the slack variable $c$ over the control parameter $\theta$, to guide the exploration of new $\theta$. 
Our framework is \emph{end-to-end fully differentiable} by the gradients from the value function in upper MBRL and from the certificates in lower SDP and LP and can be viewed as a \emph{`closed-loop} learning-verification paradigm where the failed verification provides immediate gradient feedback to the controller synthesis. We conducted experiments on linear and non-linear systems with linear and non-linear controller synthesis, demonstrating significant advantages of our approach over model-based stochastic-value gradient (SVG) and its extension to solve CMDPs for safety. Our approach can find certificates for safety and stability in most cases based on identified dynamics and provide better results of those two properties in simulations.

In the rest of the paper, Section~\ref{sec:related work} discusses related works. Section~\ref{sec:problem formulation} formulates the problem to address. Section~\ref{sec: approach} presents our proposed approach, and Section~\ref{sec:experiment} shows the experimental results. Section~\ref{sec:conclusion} concludes the paper.   

\section{Related Works}\label{sec:related work}

Certificate-based verification in our work is related to the literature on  barrier function safety~\cite{prajna2004safety} and Lyapunov stability~\cite{lyapunov1992general}, which provide formal guarantees on the safe control of systems to avoid unsafe states and on the system stability around an equilibrium point, respectively. In classical control, finding barrier or Lyapunov functions is challenging~\cite{papachristodoulou2002construction} and often requires considerable expertise and manual effort~\cite{chen2020learning} through optimization. 
Our approach, in contrast, automatically searches for certificates and provides gradient feedback from the failed searches to the learning process
, to guide the exploration of control parameters for increasing the chance of finding feasible certificates.


Regarding safety in particular, our work aims at addressing \emph{hard} safety constraints, by ensuring that the system never enters an unsafe region with formal guarantees~\cite{huang2022polar,Huang_EMSOFT19,Fan_ATVA20}, both during and after training. In contrast, in the popular CMDP-based methods~\cite{thananjeyan2021recovery,srinivasan2020learning,stooke2020responsive}, the agent aims to maximize the expected cumulative reward while restricting the expectation of cumulative cost for their unsafe interactions with
the environment under a certain threshold. 
Since the agent can still take unsafe actions with some cost, the safety constraints in CMDPs can be regarded as addressing \emph{soft} safety constraints without formal guarantees. There are also works that ensure safety by addressing stability~\cite{berkenkamp2017safe} in RL, but we consider these two properties as different in this work, where safety is defined based on the reachability of the system state. 




In terms of optimization techniques, our work leverages SVG~\cite{heess2015learning} in MBRL and is a first-order, end-to-end differentiable approach with the computation of the analytic gradient of the RL value function. Our work also conducts convex optimization for the certification. As such an optimization problem may not be feasible to solve, our approach tries to `repair' it via a  slack variable, which is differentiable to control parameters. 
This is related to but different from the approach in~\cite{barratt2021automatic}, which tries to repair the infeasible problems by modifying program parameters. Finally, as a differentiable framework, our approach is related to safe PDP~\cite{jin2021safe}, which, different from ours, requires an explicit dynamical model with no unknown parameters and an initial safe policy. 


Our work is also related to the joint learning of controller and verification by leveraging neural networks to represent certificates~\cite{jin2020neural, qin2021learning, luo2021learning, tsukamoto2020neural, robey2020learning, lindemann2021learning, dawson2022corl,chang2019neural, robey2021learning, wang2022enforcing}. These approaches first translate the certificate conditions into loss functions, sample and label data points from the system state, and then learn the certificate in a supervised learning manner. However, they require a known system dynamics model or safe demonstration data, and cannot be directly applied to systems models with unknown parameters or without safe data, which is the case our approach addresses via the RL process. 
Moreover, the neural network-based certificates generated by these approaches are often tested/validated through sampling-based methods and are not formally verified, while our approach provides formal and deterministic guarantees once the certificate is successfully obtained. 
\section{Problem Formulation}\label{sec:problem formulation}
We consider a continuous CPS whose dynamics can be expressed as an ordinary differential equation (ODE):
\begin{equation}\label{eq:sys_dynamics}
\dot{x} = f(x, u;\alpha),
\end{equation}
where $x \in X \subset \mathbf{R}^n$ is a vector denoting the system state within the state space $X$ and $u \in U \subset \mathbf{R}^m$ is the control input variable. $f:\mathbf{R}^n \times \mathbf{R}^m \rightarrow \mathbf{R}^n$ is a locally Lipschitz-continuous function ensuring that there exists a unique solution for the system ODE. Without loss of generality, $f$ is a polynomial function, as common elementary functions, such as $\sin(x), \cos(x), \sqrt{x}, \frac{1}{x}, e^{x}, \ln(x)$ and their combinations, can be equivalently transformed to polynomials \cite{liu2015abstraction}. $\alpha \in \mathbf{R}^{|\alpha|} \in [\underline{\alpha}, \bar{\alpha}]$ is a vector denoting the unknown system model parameters, which are within a lower bound $\underline{\alpha}$ and an upper bound $\bar{\alpha}$.  The system has an initial state set $X_0 \subset X$, an unsafe state set $X_u \subset X$, and a goal state set $X_g \subset X$. Without loss of generality, $X_g$ is assumed as the origin point in this paper. These sets are semi-algebraic, which can be expressed as:
$X_0 = \{ x | \xi_i(x) \geq 0, i = 1, \cdots, s \},
    X_u = \{ x | \zeta_i(x) \geq 0, i = s+1, \cdots, s+q \}, 
     X = \{ x | \psi_i(x) \geq 0, i = s+q+1, \cdots, s+q+r \}.$
The partial \textit{derivatives} $f_x$ and $f_u$ can be computed with the parameters $\alpha$. We abbreviate partial differentiation or gradient using subscripts, e.g, $\frac{\partial f(x, u;\alpha)}{\partial x} \overset{\Delta}{=}  f_x$ with \textit{gradient} or \textit{derivative} in front. 

Such a continuous system can be controlled by a feedback deterministic controller $\pi(x;\theta) = \theta \cdot w^\theta(x): X \rightarrow U$, which is parameterized by vector $\theta \in \mathbb{R}^{|\theta|}$ and monomial basis $w^\theta(x)$. Note that  $w^\theta(x)$ can contains non-linear terms.  Given any time $\forall t \geq 0$, the controller $\pi$ reads the system state $x(t)$ at $t$, and computes the control input as $u = \pi(x(t);\theta)$. Overall, the system evolves by following $\dot{x} = f(x, \pi(x;\theta))$ with $\pi$. 


A \textit{flow} function $\varphi(x(0), t): X_0 \times \mathbb{R}_{+} \rightarrow X$ maps any initial state $x(0)$ to the system state $\varphi(x(0), t)$ at time $t, \forall t \geq 0$. Mathematically, $\varphi$ satisfies 1) $\varphi(x(0), 0) = x(0)$, and 2) $\varphi$ is the solution of the $\dot{x} = f(x, u)$.
 Based on the flow definition, the system safety and stability properties and their corresponding certificates, e.g., barrier function and Lyapunov function, are defined as follows.

\begin{definition}
(\textbf{Infinite-time Safety Property}) Starting from any initial state $x(0) \in X_0$, the system defined in~\eqref{eq:sys_dynamics} is considered as meeting the safety property if and only if its \textit{flow} never enters into the unsafe set $X_{u}$:
$\forall   t\geq 0, \ \varphi (x(0), t) \notin  X_{u}.$
\label{Def:infinite-time-safety}
\end{definition}
This safety property can be formally guaranteed if the controller $\pi$ can obtain a barrier function as:
\begin{definition} \label{def:barrier}
(\textbf{Exponential Condition based Barrier Function~\cite{kong2013exponential}}) Given a controller $\pi(x;\theta)$, $B(x; \beta^B)$ is a safety barrier function parameterized by vector $\beta^B \in \mathbb{R}^{|\beta|}$ with $\lambda \in \mathbb{R}$ if:
\begin{displaymath}
    \begin{aligned}
        &B(x;\beta^B) \leq 0, \forall x \in X_0, \\
        & B(x;\beta^B) > 0, \forall x \in X_u, \\
        &\pdv{B}{x}\cdot f(x,\pi(x);\alpha) - \lambda B(x;\beta^B) \leq 0, \forall x \in X.\\
    \end{aligned}
\end{displaymath}
\end{definition}

\begin{remark}
\textit{\textbf{(Shielding-based One-Step Safety)}} Another possible way to ensure safety is to check during run-time a pre-defined shield for the system and stop the system when finding a hazard affront. Note that such shielding mechanism is reactional. It tries to protect the system from danger by only looking one step forward, which may degrade the overall performance, as shown in the experiments. Moreover, the system could still be led towards the unsafe region after several steps when taking the current action and has to be stopped eventually. In contrast, if can be found, a barrier function guarantees infinite-time safety.  \label{remark:shielding} 
\end{remark}

\begin{definition}
(\textbf{Stability Property})
Starting from any initial state $x(0) \in X_0$, the system defined in~\eqref{eq:sys_dynamics} is stable around the goal set $X_g$ if there exists a $\mathcal{KL}$ function $\tau$~\cite{khalil2002nonlinear} such that for any $x(0) \in X_0$,
$\norm{\varphi(x(0), t)}_{X_g} \leq \tau(\norm{x(0)}_{X_g}, t),$
where $\norm{x}_{X_g} = \inf_{x_g \in X_g} \norm{x - x_g}$, with $\norm{\cdot}$ denoting the Euclidean distance.
\end{definition}
This stability property can be formally guaranteed if there exists a Lyapunov function for $\pi$ as:
\begin{definition}
(\textbf{Lyapunov Function}) $V(x;\beta^V) (\beta^V \in \mathbb{R}^{|\beta|})$ is a Lyapunov function of controller $\pi(x;\theta)$ if:
\begin{displaymath}
    \begin{aligned}
    & V(x;\beta^V) \geq 0, x \in X, \\
    & \pdv{V}{x}\cdot f(x, \pi(x;\theta);\alpha) \leq 0, x \in X.
    \end{aligned}
\end{displaymath}
\end{definition}

Considering the safety and stability certificates, the problem we address in this paper can be defined as a certified control learning problem:
\begin{problem}\label{problem}\textbf{(Certified Control Learning)}
Given a continuous system defined as in~\eqref{eq:sys_dynamics}, learn the unknown dynamical parameters $\alpha$ and a feedback controller $\pi(x;\theta)$ so that the system formally satisfies the safety property and/or the stability property with barrier function $B(x;\beta^B)$ and/or Lyapunov function $V(x;\beta^V)$ as certificates.
\end{problem}
\section{Our Approach for Certified Differentiable Reinforcement Learning}\label{sec: approach}

In this section, we present our \emph{certified differentiable reinforcement learning framework} to solve the Problem~\ref{problem} defined above.
We first introduce a novel bilevel optimization formulation for Problem~\ref{problem} in Section~\ref{sec:bilevel_formulation}, by treating the learning of the controller and the system model parameters as an upper-level MBRL problem and formulating the verification as a lower-level SDP or LP problem. We connect these two sub-problems with a slack variable based on certification results. We then solve the bilevel optimization problem with the Algorithm~\ref{alg} introduced in Section~\ref{sec:algorithm}, which leverages the gradients of the slack variable and the value function in RL, and includes techniques for variable transformation, safety shielding, and parameter identification.
Finally, we conduct theoretical analysis on the soundness, incompleteness and optimality of our approach in Section~\ref{sec:theory}.

\subsection{Bilevel Optimization Problem Formulation}
\label{sec:bilevel_formulation}

In this section, we introduce a novel and general bilevel optimization problem for the certified control learning framework and then its specific extension to SDP and LP. 

In general, we can formulate a constrained optimization problem for the certified control learning defined in Problem~\ref{problem} as: 
\begin{equation}\label{eq:constrained_problem}
    \begin{aligned}
    &\max_{\theta, \alpha} \mathbb{E}_{f, x(0)\in X_0}[\mathcal{V}(x(0))], \\    \text{s.t.}~~~ & \mathcal{I}^i(x ;\theta, \alpha,\beta) \geq 0, \\ 
    & \mathcal{E}^{j}(x ;\theta, \alpha,\beta) = 0.
    \end{aligned}
\end{equation}
Here, $\mathcal{V}(x(0))$ is the value function on the initial state $x(0) \in X_0$ in RL.  $\mathcal{I}^i(x; \theta, \alpha, \beta)$, $\mathcal{E}^j(x; \theta, \alpha, \beta)$ are the inequality and equality constraints encoded from barrier certificate and Lyapunov function via various relaxation techniques (such as SDP and LP that are later introduced), with $\theta \in \mathbb{R}^{|\theta|}$, $\alpha \in \mathbb{R}^{|\alpha|}$, $\beta \in \mathbb{R}^{|\beta|}$ as the vector of controller parameters, unknown system parameters, and parameters for certificates, respectively.
As RL builds on the discrete-time MDPs, we need to discretize continuous dynamics $f$ to compute $\mathbb{E}[\mathcal{V}(x(0))]$ by simulating different traces with the controller. 
 Specifically, in RL,  $\mathcal{V}(x)$ satisfies the Bellman equation as:
$\mathcal{V}(x) = r(x, \pi(x)) + \gamma \mathcal{V}'(x'),$
where $r$ is a reward function at the state-action pair $(x, \pi(x))$, encoding the desired learning goal for the controller, $x'$ is the next system state, $\mathcal{V}'$ is the value function of $x'$, and constant factor $\gamma < 1$.

Such a constrained optimization problem in RL is often infeasible, where the infeasible certification results cannot directly guide the control learning problem. To leverage the gradient from the verification results and form an \emph{end-to-end differentiable} framework, the above problem can be modified to our \textbf{bilevel optimization problem} by introducing a slack variable $c \in \mathbb{R}^{+}$. Specifically, the upper problem tries to solve:
\begin{align*}
   \max_{\theta, \alpha}  \mathbb{E}[\mathcal{V}(x(0);\theta, \alpha)] - \lambda (c^*(\theta, \alpha))^2 - \norm{\alpha - \alpha_0}^2, 
\end{align*} 
where $c^*(\theta, \alpha)$ is the solution to a lower-level problem:
\begin{align}
    &   \qquad \min_{\beta} c,  ~ \nonumber
      \\ \textrm{subject to}~~~~  
    & \begin{cases}
    \mathcal{I}^i(x;\theta, \alpha, \beta) + c \geq 0,  \\ 
    \mathcal{E}^{i}(x ;\theta, \alpha,\beta)  \leq c,\\
    \mathcal{E}^{i}(x ;\theta, \alpha,\beta)  \geq -c, \\
     c \geq 0,
     \end{cases}
     \label{eq:bilevel-all}
\end{align}
where $\alpha_0 \in \mathbb{R}^{|\alpha|}$ is the unknown ground truth value of the unknown system model parameter vector and needs to be estimated in learning. $\lambda \geq 0$ is a penalty multiplier. Overall, the lower-level problem tries to search for a feasible solution for the certificates while reducing the slack variable $c$. The upper-level problem tries to maximize the value function in RL, reduce the penalty from the lower-level slack variable, and learn the uncertain parameters. Once the lower optima $c^* = 0$, we can obtain a feasible solution for the original constrained problem~\eqref{eq:constrained_problem} and therefore generate a certificate.
In this way, by differentiating the lower-level problem, the \textit{gradient} $c^*_{\theta}$ of $c^*$ with respect to $\theta$ can be combined with the gradient of MBRL in the upper-level problem, and the entire bilevel optimization problem is fully differentiable. 

\subsubsection{SDP Relaxation for Bilevel Formulation}
\begin{definition}
\textbf{(Sum-of-Squares)} A polynomial $p(x)$ is a sum-of-squares (SOS) if there exist polynomials $f_1(x), f_2(x), \cdots, f_m(x)$ such that $p(x) = \sum_{i=1}^{m} f_i(x)^2.$ In such case, it is easy to get $p(x) \geq 0$.
\end{definition}
For the positivity of SOS, the three conditions in a barrier function as defined in Definition~\ref{def:barrier} can be relaxed into three SOS programmings based on the Putinar's Positivstellensatz theorem~\cite{nie2007complexity}:
\begin{displaymath}
    \begin{aligned}
        & - B(x) - \sum_{i=1}^s\sigma_i(x)\cdot \xi_i(x) \in \Sigma[x], \\
        & B(x) - \sum_{i=s+1}^{s+q}\sigma_i(x)\cdot \zeta_i(x) \in \Sigma[x], \\
        &-\pdv{B}{x}\cdot f(x,\pi(x)) + \lambda B(x) - \sum_{i=s+q+1}^{s+q+r}\sigma_i(x) \psi_i(x) \in \Sigma[x].
    \end{aligned}
\end{displaymath}
Here, $\sigma_i \in \Sigma[x] \geq 0, i=1,\cdots, s+q+r$. $\Sigma[x]$ denotes the SOS ring that consists of all the SOSs over $x$. $\xi_i(x)\geq 0$, $\zeta_i(x) \geq 0$, $\psi_i(x) \geq 0 $ are the semi-algebraic constraints on $X_0$, $X_u$, and $X$, respectively. Note that in above formulation, barrier parameter $\beta$ is the decision variable while controller parameter $\theta$ and dynamics parameter$\alpha$ are fixed.
If such SOS programmings can be solved, i.e., a feasible barrier function $B(x)$ exists, the system is proved to be always safe under the controller $\pi$ and identified $\alpha$.

Similarly, a Lyapunov function can be formulated into two SOS programmings as the following, and if a solution is obtained, the system stability can be guaranteed:
\begin{displaymath}
    \begin{aligned}
        & V(x) - \sum_{i=1}^{r} \sigma_{i}(x)
        \cdot \psi_i(x) \in \Sigma[x], \\
        & -\pdv{V}{x}\cdot f(x, \pi(x)) -   \sum_{i=r+1}^{2r} \sigma_{i}(x)\cdot \psi_i(x) \in \Sigma[x].
    \end{aligned}
\end{displaymath}

Next, we are going to show how to transform an SOS into an SDP, which is used in our framework. 
    Given a polynomial $h(x)$ in SOS with the degree bound $2D$, we have:
    \begin{displaymath}
    \begin{aligned}
        h(x; \theta, \alpha, \beta) \in \Sigma[x] \Longleftrightarrow
        &h(x;\theta, \alpha, \beta)=w(x)^TQ(\theta, \alpha, \beta)w(x), \\& Q(\theta, \alpha, \beta)\succeq 0.
        \end{aligned}
    \end{displaymath}
    Here, $w(x)=(1,x_1,\cdots ,x_n, x_1x_2, \cdots, x_n^{D})$ is a vector of monomials, and $Q$ is a $d^Q \times d^Q$ positive semi-definite matrix, where $d^Q=\binom{D+n}{D}$, called \emph{Gram matrix} of $h(x)$~\cite{prajna2002introducing}.

The problem in~\eqref{eq:constrained_problem} can then be written as:
\begin{displaymath}
\begin{aligned}
    & \max_{\theta, \alpha} \mathbb{E}[V(x(0))], \\
     \text{s.t.}~~ & h^i(x;\theta, \alpha, \beta)=w(x)^TQ^i(\theta, \alpha, \beta)w(x), \\  
    &Q^i(\theta, \alpha, \beta)\succeq 0.
\end{aligned}
\end{displaymath}
 To make $h(x) = w(x)^T Q w(x)$, we need to list all the equations for coefficients in each monomial.  Given an upper bound of degree 2D of the polynomial $h(x)$, let $a=(a_1,  a_2, \cdots, a_{n}), b=(b_1, b_2, \cdots, b_{n}), \\ d=(d_1, d_2, \cdots, d_{n}), (a_i, b_i, d_i \in \mathbb{N})$ be the $n$-dimensional vectors indicating the degree of $x=(x_1, x_2, \cdots, x_n)$. Let $h(x) = \sum_{||a||_1\leq 2D} h_{a} x_a$, where $||\cdot||_1$ is the 1-norm operator and $h_a(\theta, \alpha, \beta)$ is the coefficient of $x_a = \prod_{i=1}^{n} x_i^{a_i}$. Let $Q = {Q_{bd}}$, where $\{Q_{bd}\}$ represents the entry corresponding to $x_b$ and $x_d$ in the base vector $w(x)$. Then, by equating the coefficients for all the monomials, we have
 \begin{displaymath}
 \begin{aligned}
      & h(x;\theta, \alpha, \beta)= w(x)^TQ(\theta, \alpha, \beta)w(x)  \Longleftrightarrow \\ &\forall ||a||_1 \leq 2D, h_{a}(\theta, \alpha, \beta) = { \sum_{b+d=a}} Q_{bd}
\end{aligned}
\end{displaymath}
as the equality constraints. Along with $Q \succeq 0$, the problem in~\eqref{eq:constrained_problem} can now be written as an SDP problem:
\begin{equation}\label{eq:SDP}
\begin{aligned}
    & \max_{\theta, \alpha} \mathbb{E}[\mathcal{V}(x(0))], \\
   \textrm{s.t.}~ &  h^i_a(\theta, \alpha, \beta) = {\textstyle \sum_{b+d=a}} Q^i_{bd}, \forall ||a||_1 \leq 2D, ~ Q^i(\theta, \alpha, \beta)\succeq 0. 
\end{aligned}
\end{equation}

Therefore, the bilevel optimization problem in~\eqref{eq:bilevel-all} can be written as the following problem with SDP formulation:
\begin{align*}
   \max_{\theta, \alpha}  \mathbb{E}[\mathcal{V}(x(0);\theta, \alpha)] - \lambda (c^*(\theta, \alpha))^2 - \norm{\alpha - \alpha_0}^2, 
\end{align*} 
where $c^*(\theta, \alpha)$ is the solution to a lower-level problem:
\begin{align}
    &   \qquad\qquad\qquad   \min_{\beta} c,  ~ \nonumber
      \\ \textrm{subject to}~~~~  
    & \begin{cases}
    h^i_a(\theta, \alpha, \beta) \leq \sum_{b+d=a}Q^i_{bd} + c, \forall ||a||_1 \leq 2D,  \\ 
    h^i_a(\theta, \alpha, \beta) \geq \sum_{b+d=a}Q^i_{bd} - c, \forall ||a||_1 \leq 2D,  \\ 
     Q^i(\theta, \alpha, \beta) \succeq 0, \\
     c \geq 0,
     \end{cases}
     \label{eq:bilevel-sdp}
\end{align}
where for barrier function, $i=(1,2,3)$, and for Lyapunov function, $i=(1,2)$. 

\subsubsection{LP Relaxation for Bilevel Formulation}

In addition to SDP, to improve scalability for higher-dimensional/higher-degree systems, we introduce the Handelman Representation~\cite{sankaranarayanan2013lyapunov} to encode the bilevel optimization formulation into an LP problem.


\begin{theorem}\label{theorem:handelman}
\textbf{(Handelman)}  If $p$ is strictly positive over a compact and semi-algebric set $K = \{ x | f_j(x)  \geq 0\} (j = 1, \cdots, m)$, then $p$ can be represented as a positive linear combination of the inequalities $f_j$:
\begin{equation}
p(x) = \sum_{k=1}^{n} \lambda_{k}\prod_{j=1}^{m} f_j(x)^{n_{k,j}},  \ \  \textrm{for} \ \ \lambda_{k}>0 \ \ \textrm{and} \ \ n_{k,j}\in \mathbb{N}. \nonumber
\end{equation}
\end{theorem}

Theorem~\ref{theorem:handelman} is proved in~\cite{handelman1988representing}.
Based on it, finding the general Handelman Representation of a function $p(x)$ within a semi-algebraic set $K$ defined by inequalities $\bigwedge_{i=1}^m f_j\geq 0$ is as follows.
\begin{enumerate}
    \item Fix the demand degree $D$.
    \item Generate all possible positive power product polynomials upto degree $D$ in the form $p_a = \prod_{j=1}^{m} f_{j}^{a_j}$,  where each $a_j\in\mathbb{N}$ and $\sum_{j=1}^m a_j = \norm{a}_{1} \leq D$, and thus $p_a \geq 0$ and we have $PP = \{ p_a ~|~ \forall \norm{a}_1 \leq D \}$. 
    \item Express $p(x) = \sum_{p_a\in PP} c_s p_a$ and equate the known polynomial $p$ with the set of power products from the last step to get the linear equality constraint involving $c_s$.
    \item If $c_s$ exists, then $p(x) \geq 0$ is proved.
\end{enumerate}
According to  Theorem~\ref{theorem:handelman}, Definition~\ref{def:barrier}, and Krivine-Vasilescu-Handelman’s Positivstellensatz~\cite{lasserre2005polynomial}, barrier function can be relaxed into following three LP constraints: 
\begin{align*}
\label{eq:barrier_LP}
    B(x) = \sum_{deg(p_\xi^\delta)\leq D}\lambda_\delta p_\xi^\delta, \quad -B(x) &> \sum_{deg(p_\zeta^\omega)\leq D}\lambda_\omega p_\zeta^\omega,\\
    \pdv{B}{x}\cdot f(x,\pi(x;\theta);\alpha)-\lambda B(x) &> \sum_{deg(p_\psi^\tau)\leq D}\lambda_\tau p_\psi^\tau,\\
  \lambda_\delta, \lambda_\tau, \lambda_\omega&\geq0.
\end{align*}
Here, $p_\xi^\delta$, $p_\psi^\tau$, and $p_\zeta^\omega$ are all power products of polynomials on the initial space, state space, and unsafe space as $\xi_i(x)\geq 0$, $\zeta_i(x) \geq 0$, and $\psi_i(x) \geq 0$. 
Thus, according to Theorem~\ref{theorem:handelman}, the positivity of each barrier certificate property can be ensured from the above formulation. Similarly, the Lyapunov function can be encoded into the following equations for stability properties.
\begin{equation*}
\label{eq:lyapunov_LP}
\begin{aligned}
V(x;\beta^V)&> \sum_{deg(p_\psi^\gamma)\leq D}\lambda_\epsilon p_\psi^\gamma, \\ 
-\pdv{V}{x}\cdot f(x, \pi(x;\theta);\alpha) &> \sum_{deg(p_\psi^\gamma) \leq D}\lambda_\nu p_\psi^\gamma,\\
 \lambda_\epsilon, \lambda_\nu&\geq0.
\end{aligned}
\end{equation*}
We can conduct the same constraints generation for LP as SDP by equating the coefficients of all the possible monomials.  Let $\lambda_i p_i^a$ denote the coefficient of monomial $x_a = \prod_{i=1}^{n} x_i^{a_i}, a=(a_1, \cdots, a_n)$, and thus $h^i_a(\theta, \alpha, \beta) = \lambda_i {p_i^a}$. 
Then the bilevel optimization problem in~\eqref{eq:bilevel-all} can be written as the following problem with LP formulation:
\begin{align*}
   \max_{\theta, \alpha}  \mathbb{E}[\mathcal{V}(x(0);\theta, \alpha)] - \lambda (c^*(\theta, \alpha))^2 - \norm{\alpha - \alpha_0}^2, 
\end{align*} 
where $c^*(\theta, \alpha)$ is the solution to a lower-level problem:
\begin{align}
    &   \qquad\qquad\qquad   \min_{\beta} c,  ~ \nonumber
      \\ \textrm{subject to}~~~~  
    & \begin{cases}
    h^i_a(\theta, \alpha, \beta) \leq \lambda_i {p_i^a} + c, \forall ||a||_1 \leq D,  \\ 
    h^i_a(\theta, \alpha, \beta) \geq \lambda_i {p_i^a} - c, \forall ||a||_1 \leq D,  \\ 
     \forall \lambda_i \geq 0, \\
     c \geq 0.
     \end{cases}
     \label{eq:bilevel-lp}
\end{align}
\begin{remark}
Note that both SOS and Handelman relaxations are incomplete, meaning that it is possible that a polynomial $p(x)$ is positive but cannot be expressed by SOS or Handelman representations. 
\end{remark}

\subsection{Algorithm for Solving the Bilevel Optimization Problem}
\label{sec:algorithm}

We develop the following Algorithm~\ref{alg} to solve the bilevel optimization problem by SDP and LP. The inputs to Algorithm~\ref{alg} include the system model (with unknown parameters), the step length, a shielding set for ensuring the system safety during learning (more details below), and the form of polynomials for the certificates and the controller.
The outputs include the learned controller and its certificates (barrier and/or Lyapunov function).  There are four major modules in Algorithm~\ref{alg}, including variable transformation for system model, shielding-based safe learning, parameter identification, and gradient computation for RL and certificates, as below. 

      \begin{algorithm}
   \caption{End-to-end MBRL with Certification}
   \label{alg}
    \begin{algorithmic}[1]
       \STATE {\bfseries Input:} Nominal dynamics $\dot{x} = f(x, u; \alpha)$ and its discretized form $f^d$, with unknown system model parameters $\alpha \in [\underline{\alpha}, \bar{\alpha}]$, step length $l$, shielding set $\mathcal{S}$, barrier function form $B(x) = \beta^B \cdot w^B(x)$ ($\beta^B$ unknown), Lyapunov function form $V(x) = \beta^V \cdot w^V(x)$ ($\beta^V$ unknown), and controller form $\pi = \theta \cdot w^\theta(x)$ ($\theta$ unknown). 
       \STATE $\theta = 0$, $\beta^B = 0$, $\beta^V = 0$.
        \STATE Conduct variable transformation if the system model contains non-polynomial terms.
        \REPEAT 
       \STATE Sample trajectory with $\theta$; stop early if state $x \in \mathcal{S}$. \label{alg:line:shielding}
       \STATE $\mathcal{V'}_{x'} = 0, \mathcal{V'}_{\theta'} = 0$.
       \FOR{$t=T$ {\bfseries down to} $0$ with trajectory}
       \STATE     $\alpha \leftarrow \alpha +  \gamma \frac{\Delta x}{\delta t}$; computer \textit{gradients} $\mathcal{V}_x$, $\mathcal{V}_{\theta}$ as in Eq.~\eqref{eq:SVG}. 
       \ENDFOR
       \STATE Solve lower-level SDP or LP; compute $c^*, \beta^*$; compute \textit{gradient} $c^*_{\theta}$ as in Eq.~\eqref{eq:SDP_gradient}.
       \STATE $\theta \leftarrow \theta + l(\mathcal{V}_{\theta} - 2\lambda c^* \cdot c^*_{\theta}$), increase $\lambda$.
       \UNTIL{$c^* = 0$}
       \STATE {\bfseries Output:} $\pi(x;\theta), B(x;\beta^B), V(x;\beta^V)$.
        \end{algorithmic}
      \end{algorithm}

\smallskip
\noindent
\textbf{Variable Transformation:} 
If the system model contains non-polynomial univariate basic elementary functions such as $\sin(x), \\ \cos(x), \exp(x) , \log (x), 1/ x , \sqrt{x}$ \ or their combinations, we can \textbf{equivalently} transform them into polynomial terms with additional variables~\cite{liu2015abstraction}. For example if $\dot{x} =\sin(x)$, we can let $m = \sin(x), n = \cos(x)$, and we then have $\dot{x} = m, \dot{m} = \cos(x)\dot{x} = nm, \dot{n} = -\sin(x)\dot{x} = -m^2$ as a polynomial system.

\smallskip
\noindent
\textbf{Shielding-based Safe Learning for Training:}
 We compute a shielding set $\mathcal{S}$ to ensure system safety \emph{during learning}, by stopping the current learning process if the system is within $\mathcal{S}$ (line~\ref{alg:line:shielding} in Algorithm~\ref{alg}). Specifically, we can construct $\mathcal{S}$ offline, based on the definition that the system may enter the unsafe state set $X_u$ in the next step when it is in $\mathcal{S}$, i.e., 
    $\mathcal{S} =  \{ x | \min_{\alpha} \min_{x_u \in X_u} \norm{x' - x_u} = 0 \} \ \ 
    \textrm{s.t.}~~   \underline{\alpha} \leq \alpha \leq \bar{\alpha}$. 
Here, $x$ is the current state and $x'$ is the predicted next state based on some $\alpha \in [\underline{\alpha}, \bar{\alpha}]$ for the discretized system model $f^d$ (by applying zeroth-order hold to the continuous system model $f$).
During the learning process, when the system is within the shielding set $\mathcal{S}$, the current learning process will stop and start over again. Note that we compute the set with the entire interval of $\alpha$ and it only provides \emph{one-step safety} as explained in Remark~\ref{remark:shielding}. Also note that the shielding set is for ensuring safety \emph{during} learning. Regardless of whether we use it, the controller we obtained \emph{after} learning, if it exists with the generated barrier function based on the identified dynamics, is always guaranteed to be \emph{infinite-time safe} as defined in Definition~\ref{Def:infinite-time-safety}.


\smallskip
\noindent
\textbf{Parameter Identification for System Model:}
To learn the unknown parameters of the system model during RL, we can compute the state difference between any two adjacent control time steps and then compute the approximated gradient for parameters $\alpha$. For instance, for a one-dimensional system $\dot{x} = \alpha x$, we can perform
    $\alpha \leftarrow \alpha +  \gamma  \cdot [  f^d(x(\delta t),\pi(x(\delta t));\alpha) - x(\delta t)] / \delta  $,  
where $\gamma$ is the learning rate and $f^d$ is the discretized system model from the continuous system model $f$, as long as the sampling period $\delta$ is small enough according to the Nyquist–Shannon sampling theorem. In the experiments, we observe that $\alpha$ always converges to its ground truth with the learning (i.e., $\norm{\alpha - \alpha_0}^2 \rightarrow 0$), albeit we cannot guarantee the convergence. Note that the safety or stability guarantee is established on the identified system parameters, and we have the following remark.
 
\begin{remark}
\textbf{(Parameter Identification Error and Certification)} The certification is built on the identified system parameters. However, due to the errors from the discretization  and gradient approximation, the final identified parameters may be closed to the ground truth value but not the same (the ground truth value is in fact assumed as unknown in this paper). However, if the identification error can be quantified, it can then be viewed as a bounded disturbance to the system. In which case our approach can be easily extended to such disturbed systems as barrier functions can be built on uncertain parameters~\cite{prajna2004safety} and parametric Lyapunov function can also be synthesized for LTI systems~\cite{fu2000parametric}.
\end{remark}

Next, we introduce two gradients that can be computed end-to-end to solve the bilevel problem. 

\smallskip
\noindent
\textbf{Computing the Value Function Gradient in MBRL:}
As mentioned above, by applying zeroth-order hold to the continuous system model $f$, we can obtain the discrete-time model $x(t+1) = f^d(x(t), \pi(x(t)))$.
  Note that $f^d$ contains the unknown system parameters $\alpha$, which are updated at run-time.
By differentiating the Bellman equation $\mathcal{V}(x) = r(x, \pi(x)) + \gamma \mathcal{V}'(f^d(x, \pi(x)))$~\cite{heess2015learning}, we can obtain the value function \textit{gradients} $\mathcal{V}_{x}, \mathcal{V}_{\theta}$ with respect to the state $x$ and the controller parameters $\theta$:
\begin{equation}\label{eq:SVG}
\begin{aligned}
    &\mathcal{V}_x = r_x + r_u\pi_{x} + \gamma \mathcal{V}^{'}_{x'}(f^d_x + f^d_u\pi_{x}), \\ 
    & \mathcal{V}_{\theta} = r_u \pi_{\theta} + \gamma \mathcal{V}^{'}_{x^{'}} f^{d}_{u} \pi_{\theta} + \gamma \mathcal{V}^{'}_{\theta},
\end{aligned}
\end{equation}
where every subscript is a partial \textit{derivative}. $\mathbb{E}[\mathcal{V}(x(0))]$ will be increased by updating $\theta$ with the direction as \textit{gradient} $\mathcal{V}_{\theta}(x(0))$. For the implementation, we can collect a trajectory $\{x(0), u(0), r(0), \\ \cdots, x(T), u(T), r(T) \}$ of the discrete-time system by the controller, let $\mathcal{V'}_{x^T} = 0, \mathcal{V'}_{\theta} = 0$ and roll back to the initial state $x(0)$, and obtain the \textit{gradient} $\mathcal{V}_\theta(x(0))$ based on equation~\eqref{eq:SVG}.

\smallskip
\noindent
\textbf{Computing the Certification Gradient:}
 To solve the bilevel problem in an end-to-end manner, the slack variable $c^*$ should be differentiable to the controller parameters $\theta$ as it connects the two sub-problems. The lower-level SDP or LP belongs to the disciplined parameterized programming problem where the optimization variables are $c, \beta$ and the parameters are $\theta$. And the lower-level problem defined in~\eqref{eq:bilevel-all} can be viewed as a function mapping of $\theta$ to the optimal solution $(c^*, \beta^*)$, e.g., $\mathcal{F}: \theta \rightarrow (c^*, \beta^*)$.
According to~\cite{agrawal2019differentiable},  function $\mathcal{F}$ can be expressed as the composition $\mathcal{R} \circ s \circ \mathcal{C}$, where $\mathcal{C}$ represents the canonical mapping of $\theta$ to a cone problem $(A, e)$, which is then solved by a cone solver $s$ and returns $(\bar{c}^*, \bar{\beta}^*)$. Finally, the retriever $\mathcal{R}$ translates the cone solution $(\bar{c}^*, \bar{\beta}^*)$ to the original solution $(c^*, \beta^*)$. Thus, according to the chain rule, we have 
\begin{equation}
c^*_{\theta}  = \mathcal{R}_{(\bar{c}^*, \bar{\beta}^*)} \cdot s_{(A,  e)} \cdot \mathcal{C}_{\theta}
\label{eq:SDP_gradient}
\end{equation}
as a part of the \textit{gradient} in the upper-level objective.
Overall, the controller is updated as 
$\theta \leftarrow \theta + l(\mathcal{V}_{\theta} - 2\lambda c^* \cdot c^*_{\theta})$.
As the termination condition in Algorithm~\ref{alg}, $c^* = 0$  indicates that the  original constrained problem has a feasible solution $\beta^*$ given the current $\theta$, meaning that there exists a certificate for the learned controller. 



\subsection{Theoretical Analysis on Optimality}
\label{sec:theory}

\begin{proposition}\label{soundness}
\textbf{(Soundness)} Our approach is sound as the final learned controller is formally guaranteed to hold the barrier certificate for safety and Lyapunov function for stability, based on the identified parameters in the system model.  
\end{proposition}
Take the SDP relaxation as an example, the soundness is easy to check as the slack variables $c$ of the learned controller is reduced to $0$ and thus the solution of the bilevel optimization problem~\eqref{eq:bilevel-sdp} is a solution to problem~\eqref{eq:SDP}. And it is similar for the LP relaxation.

\begin{remark}
\textbf{(Incompleteness)} Our approach is incomplete as we cannot guarantee our approach will always be able to search a controller with a barrier certificate and Lyapunov function. This is due to 1) the incompleteness of the SDP and LP relaxation approaches we utilized, 2) the limited controller parameter space we optimize on, and 3) the gradients of RL and slack variable affect each other. 
\end{remark}

However, with some mild assumptions, we can provide the following completeness analysis for our framework. Here we take the SDP relaxation of the bilevel problem as an example, and the analysis can also be applied to the LP-based bilevel problem.

\begin{proposition}\label{prop:stationary_point}
\textbf{(Stationary Point)} Suppose that there exists a step length $l$ satisfying the Wolfe conditions~\cite{wolfe1969convergence} for the value \textit{gradient} $\mathcal{V}_{\theta_k}$ and the verification \textit{gradient} $c^*_{\theta_k}$ at the $k$-th update. Then Algorithm~\ref{alg} will reach a stationary point for problem~\eqref{eq:bilevel-sdp}.
\end{proposition}

$\mathbb{E}[\mathcal{V}(x(0);\theta, \alpha)] - \lambda (c^*(\theta, \alpha))^2$ can be viewed as an unconstrained optimization problem over $\theta$. For this problem, since we can compute its closed-form \textit{gradient} as $\mathcal{V}_{\theta}$ and $c^*_{\theta}$, we can choose the step length $l$ by the Wolfe conditions, which lead problem~\eqref{eq:bilevel-sdp} to a stationary point when given a $\lambda$. The proof of Proposition~\ref{prop:stationary_point} can be adapted from the general analysis in~\cite{nocedal1999numerical} and is shown below.

Reaching a stationary point does not necessarily mean that $c^* = 0$ in problem~\eqref{eq:bilevel-sdp} and thus does not necessarily lead to a solution of~\eqref{eq:SDP}. However, with stronger assumptions, we can guarantee $c^* = 0$ (and thus a solution of~\eqref{eq:SDP}) as follows in Theorem~\ref{theorem:penalty_method}. The proof of this theorem is also adapted from~\cite{nocedal1999numerical} and shown below.


\begin{theorem}\label{theorem:penalty_method}
\textbf{(Global Solution)} Suppose that $\theta_k$ is the exact global maximizer of the objective function in problem~\eqref{eq:bilevel-sdp} at the $k$-th update iteration, and that $\lambda_k \rightarrow \infty$. Then every limit point $\theta^*$ of the sequence $\{\theta_k\}$ is a global solution of~\eqref{eq:SDP}.
\end{theorem}


\noindent
\textbf{Proof for Proposition~\ref{prop:stationary_point}:} 
Let $g(\theta_k) = -\mathbb{E}[\mathcal{V}(x(0);\theta_k, \alpha)] + \lambda_k c^2(\theta_k)$, which is the negative value of the objective in the bilevel problem~\eqref{eq:bilevel-sdp} and needs to be minimized. Let $p_k = -g_{\theta_k}$  denote the line search direction as the gradient. According to the second Wolfe condition with two constant numbers $0< d_1 < d_2 < 1$, we have
\begin{displaymath}
(g_{\theta_{k+1}} - g_{\theta_{k}})^T p_k \geq (d_2 - 1) g_{\theta_{k}}^T p_k.
\end{displaymath}
Assume that the \textit{gradient} $g_{\theta}$ of $g(\theta)$ is Lipschitz continuous, which implies that there exists a constant value $L$ such that
\begin{displaymath}
\begin{aligned}
& \frac{g_{\theta_{k+1}} - g_{\theta_{k}}}{\theta_{k+1} - \theta_{k}}  = \frac{g_{\theta_{k+1}} - g_{\theta_{k}}}{l_k p_k}\leq L,\\
&(g_{\theta_{k+1}} - g_{\theta_{k}})^T p_k \leq l_k L\norm{p_k}^2,
\end{aligned}
\end{displaymath}
where $l_k$ is the step length. Combine the two inequalities, we have
\begin{displaymath}
l_k \geq \frac{d_2 -1}{L} \frac{g_{\theta_k}^T p_k}{\norm{p_k}^2}.
\end{displaymath}
According to the first Wolfe condition, we can obtain
\begin{displaymath}
\begin{aligned}
    &g(\theta_{k+1}) \leq g(\theta_k) - d_1 \frac{1 - d_2}{L} \frac{(g_{\theta_k}^T p_k)^2}{\norm{p_k}^2}, \\
    &g(\theta_{k+1}) \leq g(\theta_k) - d\norm{g_{\theta_k}}^2,
\end{aligned}
\end{displaymath}
where $d = \frac{d_1(1- d_2)}{L}$. We can then extend the inequality to the initial value as 
\begin{displaymath}
g(\theta_{k+1}) \leq g(\theta_0) - d\sum_{i=0}^{k} \norm{g_{\theta_i}}^2.
\end{displaymath}
Since we are considering the episodic RL and $c^*$ is bounded from the lower-level SDP, function $g$
is bounded, and thus there is a positive number $N$ such that
\begin{displaymath}
\sum_{k=0}^{\infty} \norm{g_{\theta_k}}^2 < N \Longrightarrow \lim_{k\rightarrow \infty} \norm{g_{\theta_k}} = 0,
\end{displaymath}
which indicates that the solving the problem~\eqref{eq:bilevel-sdp} with Algorithm~\ref{alg} eventually reaches a stationary point.
\hfill \qedsymbol

\bigskip
\noindent
\textbf{Proof for Theorem~\ref{theorem:penalty_method}:} 
Problem~\eqref{eq:SDP} can be viewed as a constrained problem with constraint $c(\theta) = 0$. Suppose that $\bar{\theta}$ is a global solution of problem~\eqref{eq:SDP} with $c(\bar{\theta}) = 0$ (meaning that there exists a feasible certificate), and name the objective function of problem~\eqref{eq:bilevel-sdp} $\max_{\theta, \alpha}  \mathbb{E}[\mathcal{V}(x(0);\theta, \alpha)] - \lambda (c^*(\theta, \alpha))^2 - \norm{\alpha - \alpha_0}^2$ as $g(\theta)$, we then have 
\begin{displaymath}
g(\bar{\theta}) \geq g(\theta) \ \forall \theta, c(\theta) = 0.
\end{displaymath}

Since $\theta_k$ maximizes $g(\theta, \lambda_k)$ for each iteration $k$, we then have $g(\theta_k, \lambda_k) \geq g(\bar{\theta}, \lambda_k)$, resulting in the following inequality:
\begin{displaymath}
\begin{aligned}
\mathbb{E}[\mathcal{V}(x(0);\theta_k, \alpha)] - \lambda_k c^2(\theta_k) &\geq \mathbb{E}[\mathcal{V}(x(0);\bar{\theta}, \alpha)] - \lambda_k c^2(\bar{\theta}) \\ &= \mathbb{E}[\mathcal{V}(x(0);\bar{\theta}, \alpha)],
\end{aligned}
\end{displaymath}
and thus, 
\begin{displaymath}
c^2(\theta_k) \leq \frac{\mathbb{E}[\mathcal{V}(x(0);\theta_k, \alpha)] - \mathbb{E}[\mathcal{V}(x(0);\bar{\theta}, \alpha)]}{\lambda_k}.
\end{displaymath}
Suppose that $\theta^*$ is a limit point of the sequence $\{\theta_k\}$, so that there exists an infinite sub-sequences $\mathcal{K}$ such that 
\begin{displaymath}
\lim_{k\in \mathcal{K}} \theta_k = \theta^*.
\end{displaymath}
When $k \rightarrow \infty$, we then have
\begin{displaymath}
c^2(\theta^*) = \lim_{k\in\mathcal{K}} c^2(\theta_k) \leq \lim_{k \in \mathcal{K}} \frac{\mathbb{E}[\mathcal{V}(x(0);\theta_k, \alpha)] - \mathbb{E}[\mathcal{V}(x(0);\bar{\theta}, \alpha)]}{\lambda_k}.
\end{displaymath}

For $\mathbb{E}[\mathcal{V}(x(0);\theta_k, \alpha)] $ and $\mathbb{E}[\mathcal{V}(x(0);\bar{\theta}, \alpha)]$, since they follow the same distribution on $x(0) \in X_0$, and we deal with the episodic setting in RL, their difference is bounded. Because we have $\lambda_k \rightarrow \infty$ when $k \rightarrow \infty$, so $c(\theta^*) = 0$, meaning that $\theta^*$ is the feasible solution of the problem~\eqref{eq:SDP}.

Moreover, follow the inequality of $\theta_k$ with $k \rightarrow \infty$, we have
\begin{displaymath}
\begin{aligned}
    \lim_{k\in \mathcal{K}} &\mathbb{E}[\mathcal{V}(x(0);\theta_k, \alpha)] - \lambda_k c^2(\theta_k) \geq  \mathbb{E}[\mathcal{V}(x(0);\bar{\theta},  \alpha)], \\
    & \mathbb{E}[\mathcal{V}(x(0);\theta^*, \alpha)] - \lambda_k c^2(\theta^*) \geq  \mathbb{E}[\mathcal{V}(x(0);\bar{\theta},  \alpha)], \\
    & \mathbb{E}[\mathcal{V}(x(0);\theta^*, \alpha)]  \geq  \mathbb{E}[\mathcal{V}(x(0);\bar{\theta},  \alpha)].
\end{aligned}
\end{displaymath}
Since $\theta^*$ is a feasible solution with $c(\theta^*)=0$, whose objective is not smaller than that of the global solution $\bar{\theta}$, we can conclude that $\theta^*$ is a global solution as well, as claimed in Theorem~\ref{theorem:penalty_method}.
\hfill \qedsymbol



\section{Experimental Results}\label{sec:experiment}

\noindent \textbf{Experimental Settings:}
As the focus of our work is to learn certified controllers that formally guarantee system safety and stability, we will compare our approach with an SVG($\infty$)~\cite{heess2015learning} based method over a variety of benchmarks. For fair comparison, the SVG is equipped with parameter identification and shielding \emph{during both training and testing}. For our approach, shielding is used \emph{only during training} but not needed at testing, as the safety is already guaranteed by the barrier function. For the safety property, the baseline SVG is further extended to solve a CMDP with a safety constraint on the system state's Euclidean distance to the unsafe set being greater than a threshold. This CMDP is solved by the standard penalty method for its augmented Lagrangian problem. For the stability property, we encode it in the reward function as the negative L2 norm to the origin point (target). We apply formal verification (i.e., search for barrier or Lyapunov function) at \emph{each iteration} of the SVG to check whether safety or stability can hold. 

In our comparison, we carefully select benchmarks with 2-6 states from~\cite{prajna2004safety,  brockman2016openai, chen2014design, huang2022polar}. It is worth noting that generating certificates for a dynamical system with \textit{a given controller} is already an NP-hard problem~\cite{prajna2006barrier} in theory and difficult to solve in practice. Current state-of-the-art works of learning-based controller synthesis with certificate therefore mainly focus on low-dimensional systems with fewer than 6 dimensional states~\cite{dawson2022safe, chang2019neural, zhao2020synthesizing, guo2020learning,robey2020learning,lindemann2021learning,luo2021learning}. Thus, we believe that the chosen benchmarks can well reflect the advantages and limitations of our approach. We test the examples on an Intel-i7 machine with 16GB memory.

\smallskip
\noindent
\noindent \textbf{Comparison on Safety and Stability:}
Table~\ref{table} summarizes the certification results of our approach with SDP and LP and the SVG-based method on four examples. We will
discuss each of them in details below.

\begin{table}
\caption{Certification results by our approach with SDP and LP, and the SVG-based method for four examples. $B^d$ denotes the successfully obtained barrier function with polynomial degree $d$ for safety guarantees, $L^d$ denotes the successfully obtained Lyapunov function with polynomial degree $d$ for stability guarantee. `$\times$' means there does not exist any certificate as the controller is unsafe. `$-$' means that it fails to find a certificate with degree up to $6$. We can see that our approach (both SDP-based and LP-based) succeeds in finding a certified controller in all cases while SVG cannot in most cases.}
\label{table}
\resizebox{\linewidth}{!}{
\begin{tabular}{lccccr}
\toprule
Examples & PJ & Pendulum & LK & Att. Control \\
\midrule
Ours (LP) & $B^{2}$ & $L^2$ & $L^2, B^6$ & $L^2$ \\
Ours (SDP)  & $B^{2}$ & $L^{2}$ & $L^{2}, B^{4}$ & $L^2$\\
SVG (CMDP) & $\times$ & $-$ & $L^{2}, -$ & $-$ \\


\bottomrule
\end{tabular}
}
\end{table}

\begin{figure}[tbp]
\centering
    \includegraphics[width=.85\linewidth]{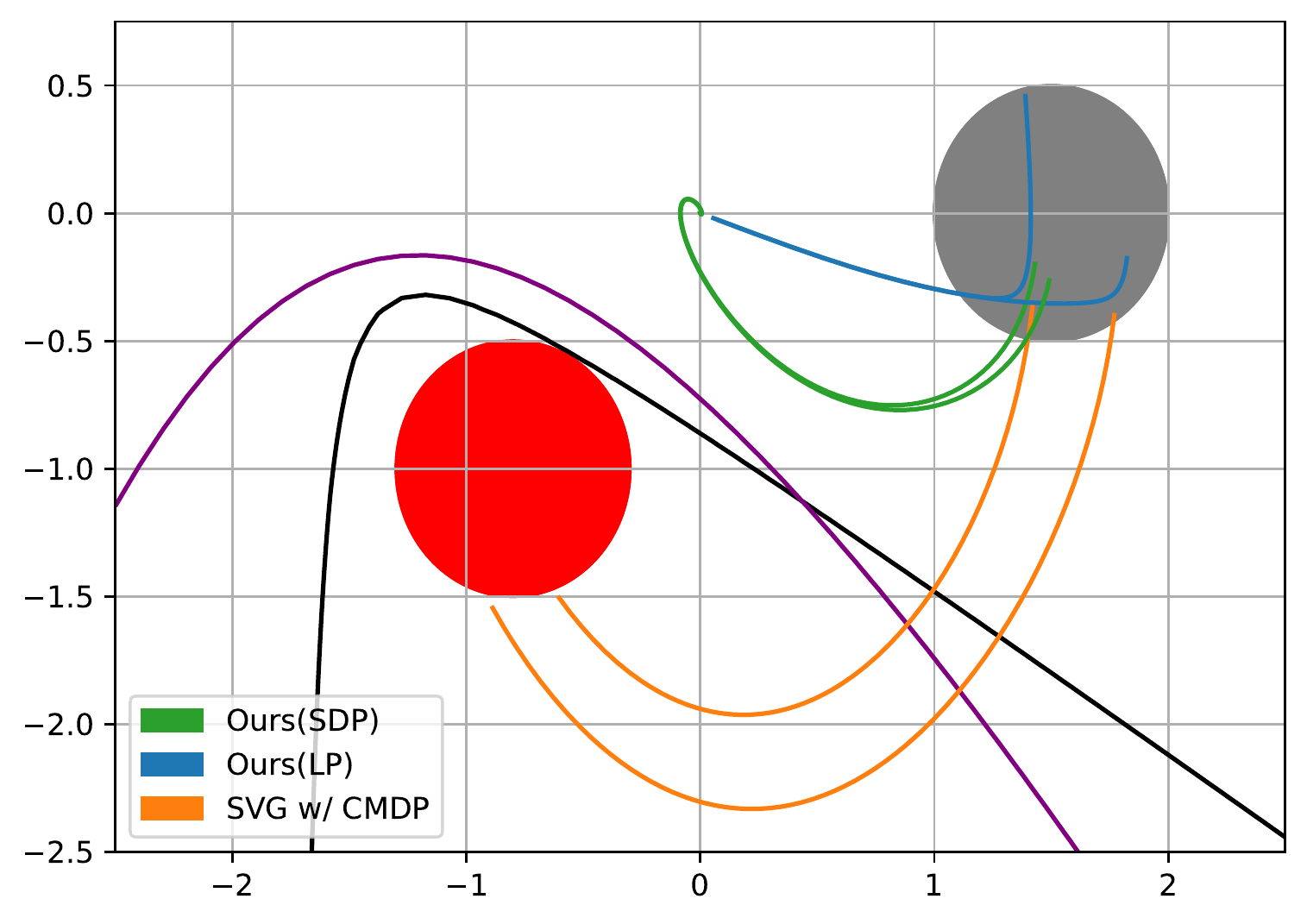}
    \caption{Trajectories under the learned controllers from our approach with SDP and LP, and from SVG (with shielding at testing) for the PJ example. Initial space $X_0$ is in grey, and unsafe region $X_u$ is in red. The barrier function 0-level plot by ours with the SDP formulation is in black and with the LP formulation is in purple.}
    \label{fig:PJ04}
\end{figure}

\textit{\underline{PJ (Safety).}}
We consider a modified example from~\cite{prajna2004safety}, whose dynamics is expressed as 
\begin{displaymath}
    \dot{x_1} = \alpha_1 x_2, \quad 
    \dot{x_2} = \alpha_2 x_1^3 + u
\end{displaymath}
where state $x_1, x_2\in [-100, 100]$, and $\alpha_1, \alpha_2 \in [-1.5, 1.5]$. The initial and unsafe sets are $X_0 = \{(x_1 - 1.5)^2 + x_2^2  \leq 0.25\}$,  $X_u = \{(x_1 + 0.8)^2 + (x_2 + 1)^2 \leq 0.25$\}. 
We focus on finding a linear controller and barrier certificate for system safety in this example.
Fig.~\ref{fig:PJ04} shows the simulated system trajectories by the learned controllers from our approach and from the SVG method with CMDP. It also shows the 0-level contour plot of the barrier functions from our approach. We can see that the controller learned by the SVG after 100 iterations is unsafe (entering the unsafe region in red and has to be stopped by shielding) and thus having no safety certificate. Our approach is safe during and after learning with shielding and the learned barrier certificate. 


\begin{figure}
     \centering
     \includegraphics[width=.9\linewidth]{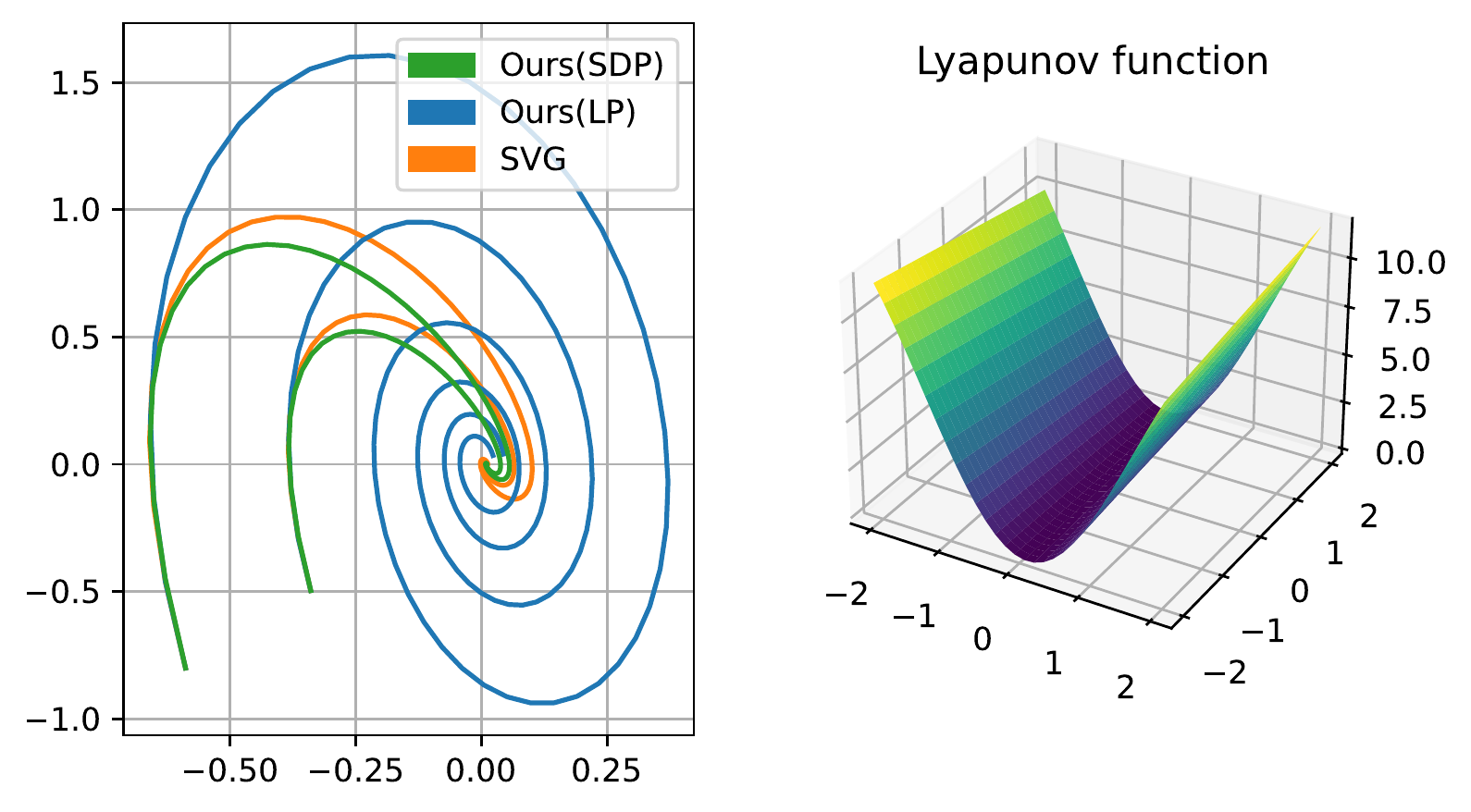}
     \caption{Trajectories on $(\varphi, \dot{\varphi})$ under the learned controllers from our approach by SDP and LP, and from baseline SVG for the Pendulum example. The right subplot shows the affiliated Lyapunov function obtained by SDP for the learned controller in our approach.}
     \label{fig:pendulum}
 \end{figure}

\textit{\underline{Inverted Pendulum (Stability).}}
We consider the inverted pendulum example from the gym environment~\cite{brockman2016openai} by a non-linear controller with $\sin{\varphi}$ and $\cos{\varphi}$ terms. The pendulum can be expressed
\begin{displaymath}
\Ddot{\varphi} = -\frac{g}{l}sin(\varphi) - \frac{d}{ml^2}\dot{\varphi} + \frac{u}{ml^2},
\end{displaymath}
where $\varphi$ is the angle deviation. 
System state $(\varphi, \dot{\varphi}) \in \{\varphi^2 + \dot{\varphi}^2 \leq 2\}$. $m = 1, l = 1, d = 0.1$. Unknown parameter $g \in [9, 10.5]$. $X_0 = \{\varphi^2 + \dot{\varphi}^2 \leq 1\}$. 
For the $\sin$ function, we conduct variable transformation with $p = \sin(\varphi)$ and $q = \cos(\varphi)$, so that the dynamics can be transformed into a 4D polynomial system. 
 Fig.~\ref{fig:pendulum} shows the trajectories from our approach (with SDP and LP) and from SVG, along with the Lyapunov function generated by our approach with SDP. 
 The SVG fails to generate a Lyapunov function with polynomial degree up to 6 during the entire learning, as shown in Table~\ref{table} while our approach succeeds with quadratic certificates by both SDP and LP.  



\begin{figure}
     \centering
     \includegraphics[width=.9\linewidth]{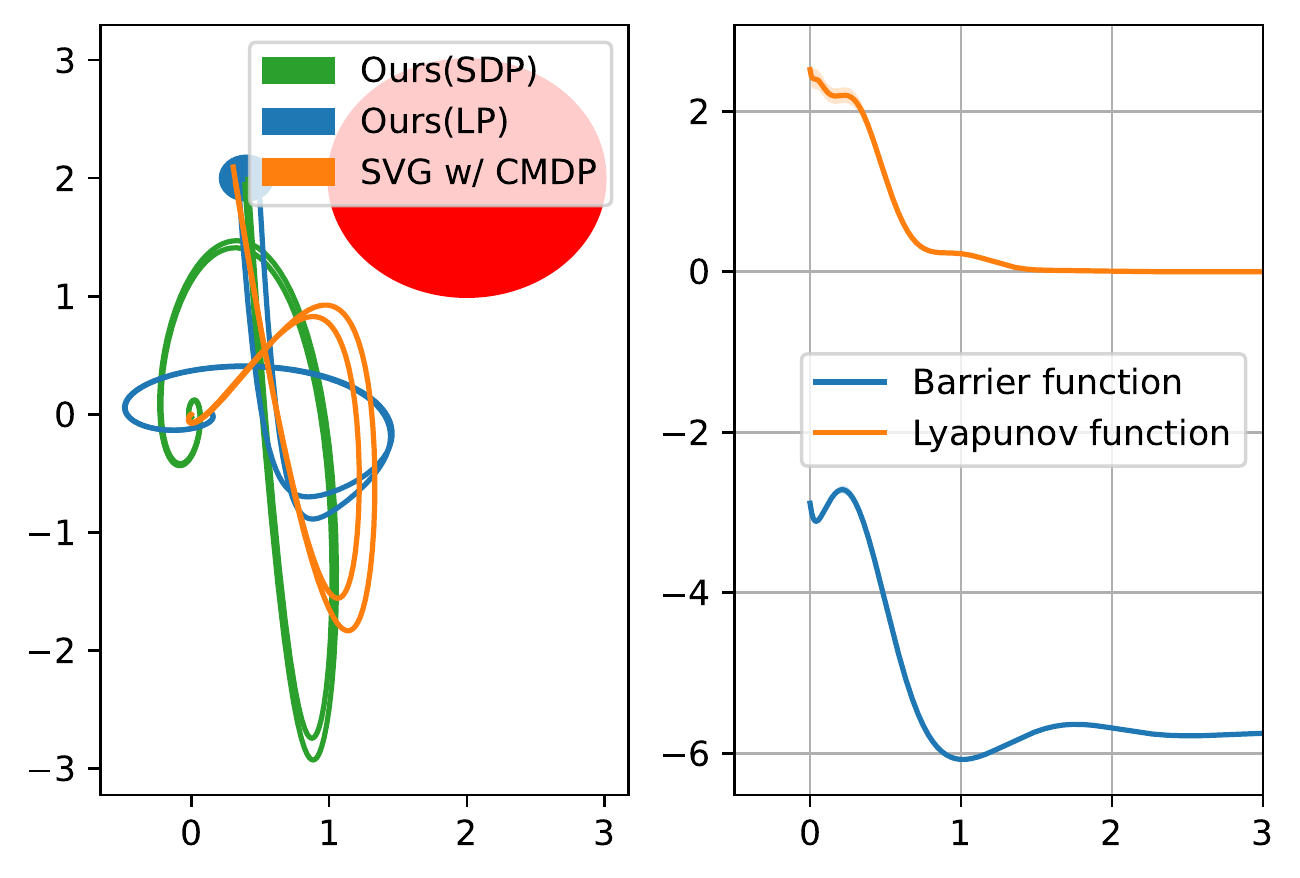}
     \caption{\textit{Left:} Trajectories on $(y, v_y)$ under the learned controllers from ours and from SVG with CMDP for Lane Keeping example. \textit{Right:} Barrier and Lyapunov function values generated by our approach with LP, along with trajectories.}
     \label{fig:LK}
\end{figure}

\textit{\underline{Lane Keeping (Safety and Stability).}}
We consider a lane keeping example~\cite{chen2014design}, where we try to derive a linear controller with barrier and Lyapunov certificate functions for both safety and stability. The system can be expressed as
$\dot{x} = Ax + Bu$, where
\begin{displaymath}
A = \begin{bmatrix}
0 & 1 & v_x & 0 \\
0 & \frac{-(C_{\alpha f} + C_{\alpha r})}{mv_x} & 0 & \frac{bC_{\alpha r} - aC_{\alpha f}}{mv_x} - v_x + \alpha_1 \\
0 & 0 & 0 & 1\\
0 & \frac{bC_{\alpha r} - aC_{\alpha f}}{I_z v_x}& 0  & \frac{-(a^2C_{\alpha f} + b^2C_{\alpha r})}{I_z v_x} + \alpha_2
\end{bmatrix},
B = \begin{bmatrix}
0 \\ \frac{C_{\alpha f}}{m} \\ 0 \\ \frac{a C_{\alpha f}}{I_z}
\end{bmatrix}.
\end{displaymath}
Here, $x = (y, v_y, \psi_e, r)^T$ is the system state with lateral displacement error $y$, lateral velocity $v_y $, yaw angle error $\psi_e$ and yaw rate $r$. Control input $u$ represents the steering angle at the front tire.  $v_x$ is the longitudinal vehicle velocity. 
$\alpha_1 \in [-15, 5], \alpha_2 \in [-10, -1]$ are the unknown parameters and other symbols are all known constants.  
$X_0 = \{\|x-x_0\|_2 \leq 0.2 \}$, $X_u = \{\| x-x_u \|_2 \leq 1 \}$, and $X = \{ \|x\|_2 \leq 3 \}$, where $x_0 = (0.4, 2, 0.5, 0)^T$ and $x_u=(2, 2, 0, 1)^T$.
Fig.~\ref{fig:LK} shows the simulated system trajectories under the controllers from our approach with SDP and LP, and from the SVG with CMDP. It also shows the barrier function value and the Lyapunov function value generated by our approach with LP, along with the trajectories over time. The SVG with CMDP can generate a quadratic Lyapunov function for stability but fails to find a barrier function for safety, as also shown in Table~\ref{table}. LP succeeds with polynomial degree $6$ for barrier function and SDP succeeds with degree $4$. Therefore, SDP takes shorter time for each iteration in this example. 

\textit{\underline{Attitude Control(Stability).}} The attitude control example~\cite{huang2022polar} is the most complex one we tested. It has 6D state and 3D control input, which can be expressed as
\begin{align*}
&\dot{\omega}_1 = \alpha_1({u_0} + {\omega_2\omega_3}), \\ 
&\dot{\omega}_2 = \alpha_2({u_1} - {3\omega_1\omega_3}), \\
&\dot{\omega}_3 = u_2 + 2\omega_1\omega_2, \\
&\dot{\psi}_1 {=} 0.5\left(\omega_2(\psi_1\psi_2 {-} {\psi_3}) {+} \omega_3(\psi_1 \psi_3 {+} {\psi_2}) {+} \omega_1({\psi_1^2} {+} {1})\right), \\
&\dot{\psi}_2 {=} 0.5\left(\omega_1(\psi_1\psi_2 {+} {\psi_3}) {+} \omega_3(\psi_2 \psi_3{-}\psi_1) {+} \omega_2({\psi_1^2} {+} {1})\right), \\
&\dot{\psi}_3 {=} 0.5\left(\omega_1(\psi_1\psi_3{-}\psi_2) {+} \omega_2(\psi_2\psi_3{+}\psi_1) {+} \omega_3({\psi_3^2} {+} {1})\right),
\end{align*}
where the state $x = (\omega, \psi)$ consists of the angular velocity vector in a body-fixed frame $\omega \in \mathbb{R}^3$ and the Rodrigues parameter vector  $\psi \in \mathbb{R}^3$. $u \in \mathbb{R}^3$ is the control torque. State space $X = \{x \ | \ ||x||_2 \leq 2\}$, unknown dynamical parameters $\alpha_1 \in [-1, 2], \alpha_2 \in [-0.5, 1.5]$. 

\begin{figure}
    \centering
    \includegraphics[width=\linewidth]{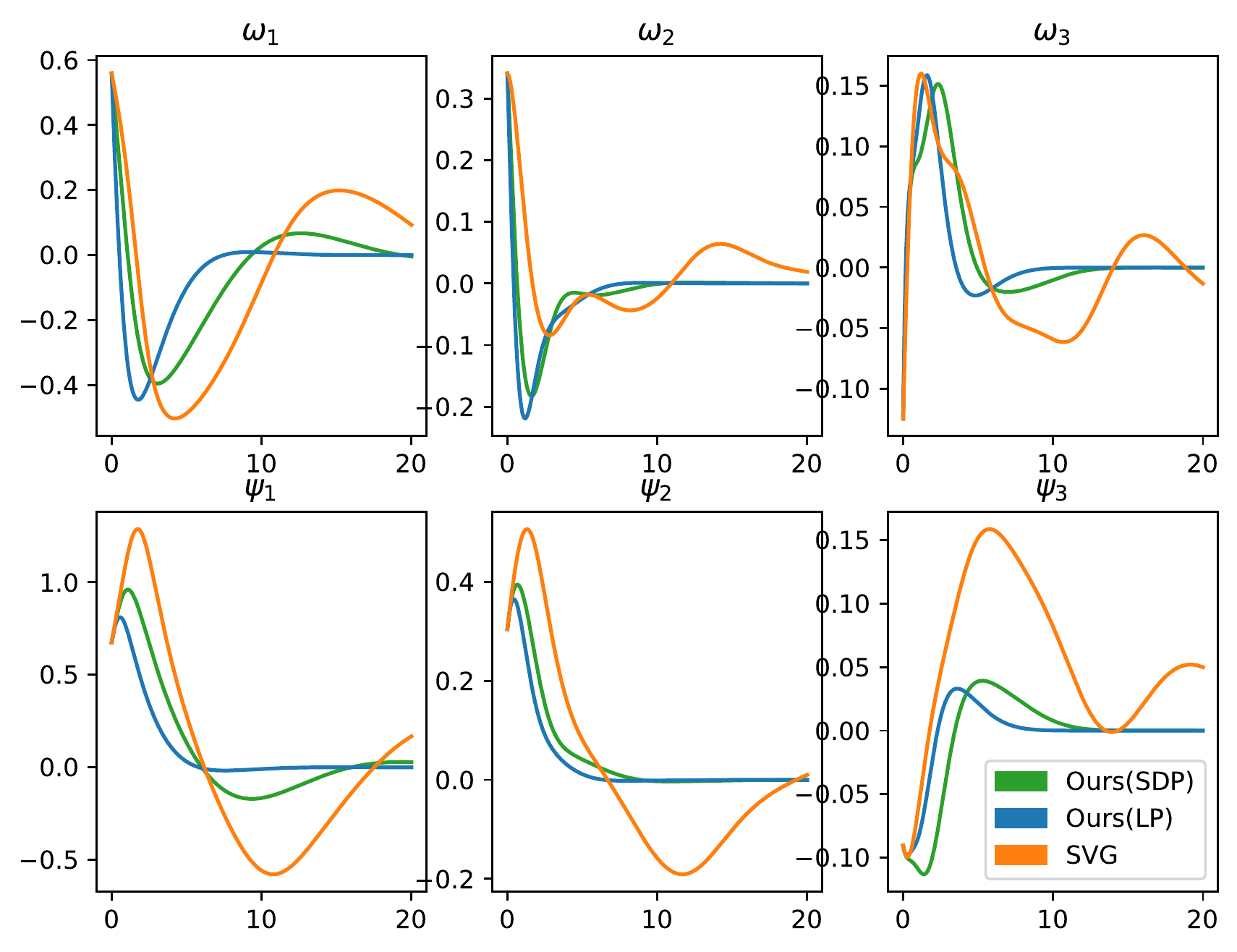}
    \caption{Trajectory on each dimension under the learned controllers from SVG and our approach for Attitude Control.}
    \label{fig:att_control}
\end{figure}

Our approach can successfully find a cubic polynomial controller with a quadratic Lyapunov function for stability with both SDP and LP, while SVG cannot generate a Lyapunov function for the entire learning process. Fig.~\ref{fig:att_control} shows the simulated trajectories by learned controllers from different approaches. 

\smallskip
\noindent
\textbf{Timing Complexity of SDP and LP Relaxation:} We first test the timing efficiency of SDP and LP in each iteration for the examples introduced before, with the polynomial degrees for certificates set as in Table~\ref{table}, and the timing results are summarized in Table~\ref{table:timing}. LP does not show a big advantage, but note that most certificate functions are quadratic in Table~\ref{table}.

\begin{figure}
    \centering
    \includegraphics[width=.93\linewidth]{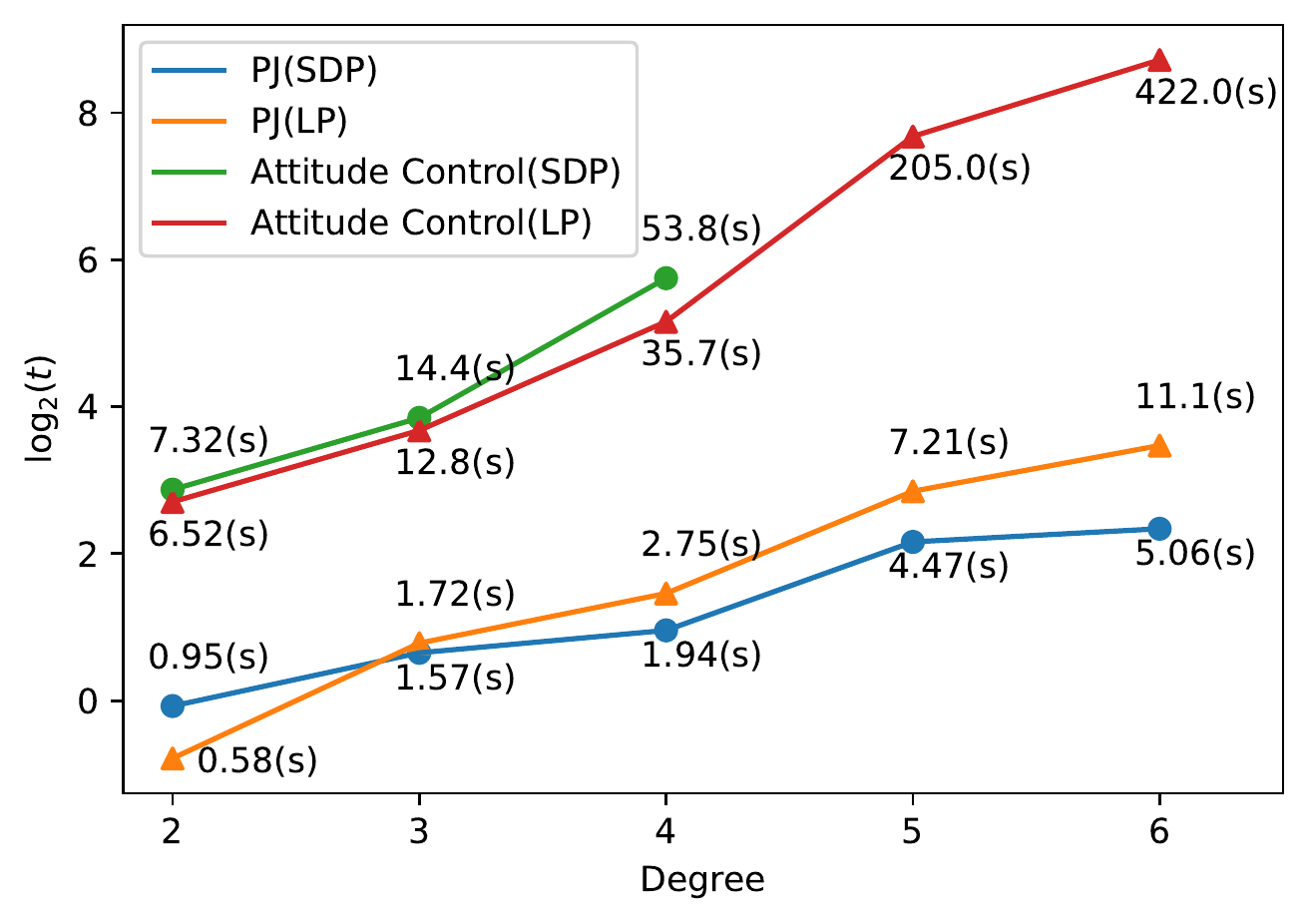}
    \caption{Averaged running time of each iteration by LP and SDP in PJ and Attitude Control examples under different degrees of certificate functions, shown in $\log_2$ magnitude with values on the plot. SDP reports timeout with degree 5 and 6 in Attitude Control, while LP can succeed. SDP is faster in the low-dimensional PJ example while LP has better scalability for the higher-dimensional systems. }
    \label{fig:timing_analysis}
    \vspace{-12pt}
\end{figure}

Thus, to further test the scalability of SDP and LP for higher-dimensional systems, we raise the degree of certificates from 2 up to 6 in testing, with runtime shown in Fig.~\ref{fig:timing_analysis}. 
We also show that the total number of variables of LP is fewer than SDP's in the Attitude Control example in Table~\ref{table:num_of_variables}, especially for higher-dimensional systems. These demonstrate LP's advantage in scalability, and we plan to explore it further in future work for larger examples.


\begin{table}
\caption{Averaged running time of each iteration by SDP and LP in our approach, with the degree shown in Table~\ref{table}. LP is typically a bit more efficient than SDP, except for generating barrier certificate in the Lane Keeping example, where SDP succeeds in polynomial degree $4$ while LP needs degree $6$. }
\resizebox{\linewidth}{!}{
\begin{tabular}{lcccr}
\toprule
 & PJ  & Pendulum & Lane Keep  &  Att. Control  \\
\midrule
SDP(s) & 0.95(B)   & 1.03(L) & 2.55(B), 0.45(L) & 7.32(L)\\
LP(s)  &  0.58(B) & 0.81(L) & 14.8(B), 0.4(L) & 6.52(L)\\
\bottomrule
\end{tabular}
}\label{table:timing}
\end{table}

\begin{table}
\caption{Number of variables in the Attitude Control example under different degree of certificate functions.}
\resizebox{0.8\linewidth}{!}{
\begin{tabular}{lccccr}
\toprule
Degree & 2  & 3 & 4  &  5 & 6 \\
\midrule
SDP & 475 & 4042 &  4168 & 26078 & 26502\\
LP  &  302 & 664 & 1380 & 2675 & 4849\\
\bottomrule
\end{tabular}
}\label{table:num_of_variables}
\vspace{-12pt}
\end{table}

\section{Conclusion}\label{sec:conclusion}
In this paper, we present a joint differentiable optimization and verification framework for certified reinforcement learning, by formulating and solving a novel bilevel optimization problem in an end-to-end differentiable manner, leveraging the gradients from both the certificates and the value function. Experimental results demonstrate the effectiveness of our approach in finding controllers with certificates for guaranteeing system safety and stability.  


\bibliographystyle{ACM-Reference-Format}
\bibliography{sample-base}




\end{document}